\pdfoutput=1

\documentclass[11pt]{article}

\usepackage[]{acl}

\usepackage{times}
\usepackage{latexsym}

\usepackage{enumitem}

\usepackage[T1]{fontenc}

\usepackage[utf8]{inputenc}

\usepackage{microtype}

\title{Towards Supporting Legal Argumentation with NLP: \\ Is More Data Really All You Need?}

\author{Santosh T.Y.S.S.$^{1}$, Kevin D. Ashley $^{2}$, Katie Atkinson$^{3}$, Matthias Grabmair$^{1}$ \\
$^{1}$Technical University of Munich, Germany\\  $^{2}$University of Pittsburgh, USA\\ $^{3}$University of Liverpool, UK}

\begin{document}
\maketitle
\begin{abstract}
Modeling legal reasoning and argumentation justifying decisions in cases has always been central to AI \& Law, yet contemporary developments in legal NLP have increasingly focused on statistically classifying legal conclusions from text. While conceptually ``simpler'', these approaches often fall short in providing usable justifications connecting to appropriate legal concepts. This paper reviews both traditional symbolic works in AI \& Law and recent advances in legal NLP, and distills possibilities of integrating expert-informed knowledge to strike a balance between scalability and explanation in symbolic vs. data-driven approaches. We identify open challenges and discuss the potential of modern NLP models and methods that integrate conceptual legal knowledge.
\end{abstract}

\section{Introduction}
Law has been an attractive domain for AI in both symbolic knowledge representation and statistical NLP. Both strands share the common goal of supporting legal practice through enhancing legal research, document analysis, drafting, and decision making. A focal question distinguishing them remains whether, and how, the process of legal reasoning \footnote{By `legal reasoning', we refer to the wide range of activities involving interpreting, arguing, and applying legal principles to reach conclusions. Legal reasoning is not a single task but a collection of related tasks around the main theme of legal decision-making as the interrelation of more of less well-defined rules and societal values with the facts of a specific case towards an outcome. Given the limited space available, we use `legal reasoning' as an umbrella term to cover the diverse contributions on this topic in the literature.} underlying all textual data shall be explicitly represented or left to opaque components, such as generative language models or neural classifiers.


In principle, legal reasoning resembles IF-THEN-like inference. Legal rules are established from sources (statutes, regulations, precedent, custom, etc.) and mandate that certain consequences follow if factual requirements are met in a specific situation. In reality, however, such logic-like inferences are interwoven with areas of ambiguity, vagueness, and human discretion \cite{urbina2002legal}. At the same time, legal orders evolve over time, continuously refining and adjusting to a dynamic world. In knowledge engineering communities, legal reasoning is characterized as `defeasible' \cite{carlos2001legal} rather than monotonic. Rules that are applicable on their face can be trumped by special exceptions, conflicting superior rules, or by distinguishing the precedent from which the rule derives. Thus legal decisions are subject to change, as they can be overturned on appeal. The evolving nature of law to align with shifting social values leads to different legal conclusions. When two parties are in conflict and desire two different resolutions, their argument will combine law and facts in a way that is beneficial to their respective goals - through adversarial discourse \cite{khairoulline2007discourse}. Legal argumentation can be seen as an exercise in competitive theory formation in front of an arbitrator, with each side constructing arguments supported by evidence, written law, cases and other authority to favor their desired conclusions while addressing pitfalls of opposing theories \cite{rissland2003ai}. 


AI \& Law as a field started started in the 1970s, when \citet{buchanan1970some} suggested that computer modeling of legal reasoning would be a promising area for research to better understand legal reasoning and argumentation.  Many approaches have been proposed over the past three decades capturing several types of reasoning by means of symbolic representations. Some 50 years after the field's beginnings, the legal profession is experiencing considerable disruption by NLP technology, most prominently large language models (LLMs). In this paper, we provide a review of AI \& Law work offering faithful modeling of legal reasoning but also requiring expensive legal expertise. We contrast this to modern, largely non-explainable, data-driven methods, which predict legal conclusions directly without engaging in any explicit legal reasoning. 

Our main contributions are as follows:
\begin{itemize}[noitemsep]
\item{An introduction to legal systems to sensitize readers to assumptions made in technical work}
\item{Surveys of (1) landmark AI \& Law work and its lessons learned, and (2) data-driven approaches to legal AI and legal NLP}
\item{A detailed discussion of perspectives to unify both strands to meet future challenges}.
\end{itemize}

Our discussion makes the following arguments:
\begin{itemize}[noitemsep]
\itemsep0em 
\item{Future work on legal AI must strive to integrate legal expertise with data-derived models}.
\item{Conveniently available legal NLP datasets come with structural assumptions, noise, and biases, which must be accounted for}.
\item{Change of legal systems over time remains an under-explored aspect in NLP works}.
\item{LLMs help alleviate knowledge acquisition bottleneck for domain model construction}.
\item{There is value in NLP that produces and assesses arguments about legal conclusions in an explainable way with domain knowledge representation}.
\item{Qualified evaluation in legal NLP is underdeveloped given the often non-well-defined nature of legal practice support tasks, resulting in exaggerated attention on convenient but uninformative benchmark metrics}.
\end{itemize}

While prior surveys of \citet{katz2023natural} and \citet{zhong2020does} focus on cataloging various use cases, tasks, and NLP techniques in legal AI, our paper critically examines the historical integration of expert knowledge into legal systems and advocates for its revival and synthesis with data-driven methods. We emphasize the unique value of expert-informed knowledge in ensuring legal reasoning aligns with established principles, which is not the primary focus of the aforementioned reviews. In contrast to \citet{mahari2023law}, which highlights the disconnect between the tasks that are pursued in legal NLP research and the actual needs of legal practitioners, our work emphasizes the critical importance of integrating expert-informed knowledge to avoid this gap. We also present directions for synthesizing expert knowledge with current technological advancements, thereby overcoming traditional bottlenecks in knowledge acquisition and enhancing the efficacy of structured argumentation models.

Most importantly, we contribute a comprehensive distillation of the conceptual ideas developed and researched by the AI \& Law community prior to the recently surging interest in law as an application domain for mainstream NLP. In part, our motivation is to connect these communities. Much legal NLP work does not build on formal models of legal knowledge and reasoning but characterizes it mostly as precursor work to modern statistical methods. Our position is that this view does not do justice to the insights gained and legal authenticity captured in this body of research. Symbolic AI \& Law has thought about how to incorporate legal expertise in models much deeper than most current NLP works, and hence the fields should merge and learn from one another. We strive to drive home the necessity of a paradigm shift in legal NLP, one that values and integrates the profound expertise of domain specialists with the capabilities of data-driven technologies.

\section{Legal Systems in a Nutshell}

Legal systems revolve around legal subjects, institutions and actors, and sources of law. While there is variation across settings, the most relevant sources typically comprise a national constitution, primary legislation (often referred to as `statutes', etc.), secondary `executive' regulation, precedents decided by courts, and other auxiliary sources. A major division exists with regard to the role of precedents relative to written law, as well as the methodology of arguing with them. Legal systems primarily influenced by continental Europe follow the `civil law' approach, where important decisions are mostly condensed into context-free interpretive rules to codified law that are compiled in secondary literature (e.g., so-called `commentaries'). In parts of the world with primarily English legal influence, so-called `common law' systems, precedents are regularly applied by means of analogizing and distinguishing arguments that take into account the facts of the case in much greater extent than civil-law-type reasoning will. International courts (e.g., the European Court of Justice, the European Court of Human Rights) usually follow hybrid methodologies that are specific to the legal regime they govern. Despite some recent diversification, virtually all AI \& Law research comes from either civil or common law backgrounds and makes corresponding assumptions, which is why we include this introduction. It is important to note that this coarse systematization is a great simplification of the world's diverse legal systems and cultures, and only intended to supplement our survey.

Written law generally consists of primary (i.e., parliamentary) legislation and secondary (i.e., executive) regulation. While enacted by two different branches of government, they are structurally similar in that they encode rules that can be formalized in IF-THEN relations. They also contain ambiguous and vague formulations in need of interpretation, for which special methods exist that are beyond the scope of this work. It is typically up to the judiciary (i.e., the courts) to settle open questions through landmark cases, often after arguments being developed in academic literature. These decisions then become part of the discourse accordingly in the applicable methodology. This transition from rule-based to case-based reasoning (in the common law) has intuitively been termed “when the rules run out” by \cite{gardner1987artificial}.


When arguing a case relative to a precedent, it is a fundamental principle of justice that similar cases should be treated alike. In common law jurisdictions, this principle is formalised in the doctrine of \textit{stare decisis}, which obliges decisions of the appropriate status to be followed when deciding a new case. Civil law legal orders also recognize a binding effect of high court precedent, but argue with them differently. 
While higher court decisions bind lower courts, cases move in the opposite direction. They are first filed in, for example, district or trial courts, where evidence is heard and first decisions are made. Decisions can then be appealed to the Appeals Courts, and eventually to Supreme Courts. At some point in this progression, arguments on evidence will be considered settled and only purely legal errors will be permissible grounds for further escalation. In a legal system, such `appeals tracks' exist for various jurisdictions (civil, criminal, administrative, etc.) and can be spread across geographic entities (e.g., federal vs state courts).

It is worthwhile to acknowledge that legal systems are inherently human-centric, involving complex decision-making processes where persuasion, interpretation, and subjective judgment play critical roles. Legal decisions are not solely about determining which side should win as a matter of justice, but about who can present the most convincing argument within the framework of established laws, principles, and precedents. The main vision of AI \& Law is that state and private actors in all aspects of the legal system can benefit from supporting software that seamlessly connects to the concepts and concerns they have been trained for and work with. Notably, legal reasoning not only happens in courts, but also in public administration and law enforcement (i.e., the executive branch), where law needs to be applied to specific situations (e.g., permits, taxes, public safety, etc.). Human accountability is paramount for the trust in the overall workings of a democratically governed society. Hence, this vision is one of AI supporting human decision makers and not replacing or unduly influencing it.

\section{Knowledge-based Approaches}
AI \& Law research started with modeling of legal reasoning by means of knowledge representation.

\noindent \textbf{Rule-based Approaches} Early landmark work demonstrated how British immigration law could be represented in Prolog \cite{sergot1986british} and outlined challenges faced in this process, including the law's rule-exception pattern, negation-as-failure (i.e., failure to prove true) vs. classical negation (logical, certain falseness), and counterfactual reasoning. \citet{waterman1980rule} developed a specialized language for rule-based legal inference. Rules establish conclusions from antecedents in a forward/backward chaining manner, thereby spanning open a derivation tree of a case outcome. 
They justify a position and explain how a conclusion can be reached, but they do not capture the dialectical aspects associated with argumentation, since no conflicting arguments are generated and no indeterminacy is accounted for. \citet{gardner1987artificial} extended by using augmented transition networks to model contract formation over time given agent actions with a basic form of uncertainty - If a condition was a `hard question' and could not be decided, the network would fork into two alternative ways to legally treat the facts. Overall, early rule-based systems were still predominantly derivations rather than argumentation models, although they correspond well to how lawyers analyze cases.

\noindent \textbf{Case-based Approaches} The adversarial nature of law naturally demands to represent arguments for both  dispute sides. The precedent-focused nature of US common law was a suitable domain for the development of what became known as `legal case based reasoning' systems. In the prominent TAXMAN system, \citet{mccarty90reflections} modeled the majority's and minority's theories and arguments  in the famous tax law case of \emph{Eisner v Macomber, 252 U.S. 189 (1920)} \cite{eisner_v_macomber_1920}, surveying the intricacies one must account for, if resolved to capture all decision-relevant concerns in depth. The HYPO system \cite{ashley1991reasoning} modeled parts of US Trade Secrets Law by means of dimensions. These are typical fact patterns that favor different sides of the dispute, and can be used to analogize and distinguish cases argumentatively by means of set comparison.  
The focal concept here is a `three-ply argument': A proponent cites the most-on-point precedent with the greatest factor overlap. The opponent distinguishes by pointing to a disfavorable factor in the precedent but not in the new case, or a favorable factor in the current case but not in the precedent, and cites a counterexample precedent. Finally, the proponent offers a rebuttal by distinguishing the counterexample. This was built upon in the CATO system \cite{aleven1997teaching}, which arranged `factors' into a hierarchy, on the basis of which more sophisticated argumentation was possible (e.g., using hierarchy parent factors).


\noindent \textbf{Hybrid \& Extended Systems} CABARET \cite{rissland1991cabaret} first combined rule-based reasoning with HYPO-style case based reasoning around ill-defined terms contained in the rules. The integration is performed via a collection of control heuristics that interleave arguments of both kinds to support a particular conclusion. 
GREBE \cite{branting1991building} further extends this hybrid architecture with formalized domain knowledge and a semantic network representation to retrieve and compare cases.
BankXX \cite{rissland1996bankxx,rissland1997evaluating} embeds HYPO-style factor-based reasoning with a domain model into a `legal theory space' that can be searched for plausible arguments.

\noindent \textbf{Integration with Prediction}
CATO had been developed as a tutoring system and did not predict case outcomes. Issue-Based Prediction (IBP) \cite{bruninghaus2003predicting} extended the factor-based representation with a model of legal `issues', each of which could be predicted via case-based reasoning. \citet{ashley2009automatically} even proposed SMILE + IBP, classifying the presence of factors in cases by means of NLP, whereas prior factor-based systems had all relied on manual factor coding of cases. It pioneered data-driven approaches for ascribing factors to be used in conjunction with a domain model without circumventing the reasoning process entirely. 

\noindent \textbf{Values, Time, and Procedure:} Berman \& Hafner explored deeper aspects of representing cases, many of which remain challenging to this day. \citealt{berman1993representing} proposed to supplement each factor with “legal purpose(s) which it affects, and each legal purpose in turn specifies whether it favours the plaintiff or defendant”. 
Parties may offer competing arguments based on factor-based case analogies. \textit{Teleological} knowledge allows a model to go beyond factual similarities to include broader jurisprudential concepts. This was highly influential in subsequent work \cite{greenwood2003towards,chorley2005empirical,DBLP:conf/jurix/WynerBA07,grabmair2011facilitating,muthuri2017compliance,grabmair2017predicting,maranhao2021dynamic}, which converged towards speaking of ``values'' rather than purposes.

\citet{berman1995understanding} contributed a pioneering model of the temporal dynamics of case-based legal reasoning: “legal precedents are embedded in a temporal context of evolving legal doctrine, which can result in a strong precedent becoming weaker over time, to the point where a skillful attorney could reasonably predict that it will no longer be followed." 
This temporal dimension has also received attention in other works \cite{rissland2011catching,henderson2019describing,prakken1998modelling,branting1993computational}.

\citealt{berman1991incorporating} observe that the support of a precedent decision for a case to be argued is linked to its respective procedural setting. They distinguish the pleading,
pre-verdict, and 
verdict stage. 
A further difference exists between decisions on procedural matters and decisions on matters of fact and/or law. A decision in favour of the defendant party based on a procedural matter (e.g., lack of evidence) may not support the same decision in a new case which shares the factual features of the precedent but is to be decided on its merits. 
The question of decision context has received limited attention in subsequent works (e.g., \citealt{wyner2009modelling,verheij2016formalizing}). Even in the recent works on NLP-based legal judgment prediction, case outcomes are often greatly simplified, up to the point of an impoverished binary variable of whether a party won the case or not.

\noindent \textbf{Theory Construction Approach:}  As \citealt{mccarty1995implementation} pointed out, “[T]he task for a lawyer or a judge in a hard case is to construct a theory of the disputed rules that produces the desired legal result, and then to persuade the relevant audience that this theory is preferable to any theories offered by an opponent”. 
\citet{bench2000using,bench2003model,bench2001theory} model a `theory' as a set of factor-based rules and preferences among them derived from value preferences. Different theories can be compared with reference to the number of cases whose outcome they explain. The rules and preference relations form tradeoffs between sets of values raised by factors in the cases. These establish preferences among rules, which in turn predict case outcomes. The CATE system \cite{chorley2004support} enabled manual creation and testing of theories as prolog programs. The AGATHA system \cite{chorley2005agatha} constructed theories autonomously using A* search. 

\noindent \textbf{Computational Argumentation:} 
Producing an argument by using a rule-driven strategy implemented with case-driven argument moves remains a central way of justifying conclusions in cases. In the 90s this was mainly pursued using dialogue games which were designed to allow an adversarial discussion between the two parties, one represented by the computer and one by the user. Examples include \citealt{gordon1993pleadings,hage1993hard,prakken1997dialectical,prakken1998modelling,loui1995rationales}. While many of the systems referenced thus far model argumentation ad hoc, the AI \& Law field interacted considerably with its neighboring discipline of general computational models of argumentation. Of particular interest in this context is the concept of `argument schemes' as well as the connection to models of so-called `abstract argumentation'.


\noindent \emph{Argument Schemes:} An argument scheme is a stereotypical pattern of reasoning primarily constituting  a claim, a set of positive premises, and, optionally, a set of negative exceptions. 
Argument schemes have a long history, as laid out in \cite{macagno2017argumentation}. 
In modern times, schemes were used by \citealt{perelman1969new} and \citealt{toulmin1958uses}. In AI \& Law, the Toulmin argument model had been historically popular. 
It recognizes different roles of statements in an argument: Claim, Qualifier/Strength, Data/Premises, Warrant/Inference, Backing, and Rebuttal. This is suitable for legal reasoning by incorporating authority for the warrant and by including a rebuttal component in recognition of the defeasible nature of legal reasoning.
\citet{walton1996argumentation} introduced a variety of schemes into AI \& Law (e.g., from Expert Opinion, from Negative Consequences, from Rules, etc). \citet{verheij2001legal,gordon2009legal} supplemented them further (e.g., from position to know, from ontology, from cases, from testimonial evidence). Schemes have become central in AI \& Law research, being used in reasoning with evidence \cite{bex2003towards,bex2011arguments}, reasoning with cases \cite{prakken2015formalization}, e-democracy \cite{atkinson2006parmenides}, statutory interpretation \cite{araszkiewicz2021critical}, and value-based argumentation \cite{grabmair2016modeling,greenwood2003towards}. 
 

\emph{Abstract Argumentation Framework:} In seminal work, \citet{dung1995acceptability} defined \textit{abstract argumentation frameworks} (AAFs), which were introduced to AI \& Law by \citealt{prakken1995logic}. An abstract argumentation framework comprises a set of arguments and set of attack relations between them. The justified arguments are then evaluated based on subsets of arguments (`extensions') defined under a range of semantics. 
The abstract nature of Dung's theory says nothing about the structure of arguments, the nature of attack or defeat, or use of preferences. This opacity, and the coupling with argument schemes, motivated the development of structured argument models. 
For example, ASPIC \cite{caminada2007evaluation} adopts an intermediate level of abstraction 
by making some minimal assumptions on the nature of the logical language and the inference rules, and then providing abstract accounts of the structure of arguments, the nature of attack, and the use of preferences. \citealt{prakken2010abstract,modgil2009reasoning} generalised the ASPIC framework to develop ASPIC+, which can capture a broader range of systems with various assumption-based argumentation and systems using argument schemes. ASPIC+ has been applied to study legal reasoning in the works of \citealt{prakken2012reconstructing, prakken2015formalization}.


Abstract Dialectical Frameworks (ADFs) \cite{brewka2013abstract} generalize the AAF representation to node-and-directed-relations form with a set of local acceptance conditions. This allows both attack and support influence, resulting in an abstract yet intuitive model for legal reasoning. 
For example, the ANGELIC method \cite{al2016methodology} uses ADFs for representing case law in an explainable inference model on the basis of a hierarchical factor representation. The maintainability of such a representation is discussed in \cite{al2016accommodating}. \citet{atkinson2019implementing} extended  to reasoning about factors with magnitude, thereby going beyond purely boolean proposition representations of cases. 


Overall, the advantages of knowledge-based approaches are that they explicitly model legal reasoning and provide explanations of inferences.

\section{Data-driven Approaches}
\textbf{Relationship to Political Science Research:}
Data originating in the legal system has been the subject of extensive analytical study in the field of \textit{empirical legal studies}, including court and judge decision/voting behavior (e.g., \citealt{segal1984predicting,kort1957predicting,nagel1963applying, ruger2004supreme}). As most of them neither model legal reasoning nor apply NLP techniques, we do not include them in our survey.

\noindent \textbf{Early AI \& Law:}
Knowledge-centered approaches can achieve high degrees of faithfulness in their representation and  explainability in their inferences, but face the `knowledge acquisition bottleneck', as they require large amounts of expertise and modeling effort. This is in contrast to data-driven models with less hand-crafted expertise. Early works by \citet{mackaay1974predicting}  used 
nearest-neighbor methods for outcome classification. In the 1990's, \citealt{
pannu1995using,bochereau1991extracting,phlipps1989neural,bench1993neural} trained neural networks to predict outcomes and derive input feature weights.  Unsurprisingly, such early applications of ML attracted criticism \cite{aikenhead1996uses, hunter1994looking}. Obtaining substantial amounts of processable data was challenging and extensive feature engineering was necessary. 
These works focused on the application of neural networks to identify how influential certain information is for the decision and did not engage in comparative benchmarking. 

\noindent \textbf{Towards Modern Legal NLP:}
Recent years saw a resurging interest in case prediction through the use of data-driven methods learning from the large datasets now available from different jurisdictions, such as the ECtHR \cite{chalkidis2019neural, chalkidis2022lexglue, chalkidis2021paragraph,aletras2016predicting, medvedeva2021automatic, says2020prediction,tyss2023leveraging,tyss2023zero,santosh2024incorporating,liu2017predictive, medvedeva2020using, says2020prediction}
Chinese Criminal Courts \cite{luo2017learning, yue2021neurjudge, zhong2020iteratively, zhong2018legal, yang2019legal},
, US Supreme Court \cite{katz2017general,kaufman2019improving}, Indian Courts \cite{malik2021ildc,shaikh2020predicting}
French court of Cassation \cite{csulea2017predicting, csulea2017exploring,bertalan2020predicting} 
Supreme Court of Switzerland \cite{niklaus2021swiss}, 
Turkish Constitutional court \cite{sert2021using},
UK courts \cite{strickson2020legal}, German courts \cite{waltl2017predicting}, Brazilian courts \cite{lage2022predicting} and Philippine courts \cite{virtucio2018predicting}.

Earlier works employed bag-of-words features \cite{aletras2016predicting, csulea2017exploring,csulea2017predicting, virtucio2018predicting, shaikh2020predicting, medvedeva2020using}. More recent approaches use deep learning techniques \cite{zhong2018legal, zhong2020iteratively, yang2019legal} involving convolutional or recurrent networks followed by adoption of pre-trained transformer models \cite{chalkidis2019neural,niklaus2021swiss}, including legal-domain specific pre-trained variants \cite{zheng2021does,chalkidis2020legal,chalkidis2023lexfiles,douka2021juribert,masala2021jurbert,xiao2021lawformer,hwang2022multi,niklaus2023multilegalpile}.
Classification tasks on legal text interrelate, and so other words have leveraged dependencies between tasks for improving models \cite{santosh2023leveraging,yue2021neurjudge,valvoda2023role,zhong2018legal,feng2022legal,ma2021legal,dong2021legal,yang2019legal,huang2021dependency,hu2018few}
and added additional loss constraints (such as contrastive learning exploiting label information), \cite{tyss2023leveraging,zhang2023contrastive,gan2022exploiting,liu2022augmenting}
and injected legal knowledge \cite{liu2023ml,santosh2023zero,santosh2024incorporating,gan2021judgment,zhong2020iteratively,feng2022legal}

Overall, one can observe a trend towards applying NLP models to legal text with little to no architectural bias or explicit domain representation. These are then compared along quantitative metrics, typically with regard to high level classification/prediction goals (e.g., case outcome variables and document-level keywords) at the cost of interpretability. As Berman \& Hafner have observed in the 1990s, however, case outcomes are highly contextual in time, procedure, and socio-legal purpose. Classification benchmarks risk decoupling a sense of technical progress towards a notion of model `understanding' from supporting a realistic task (e.g., legal argumentation) by focusing on a highly reductive representation of its outcome. For instance, case outcome predictions are often treated as binary targets based on the majority opinion, even though judges on the same bench frequently have conflicting reasoning, leading to dissenting or concurrent opinions \cite{xu2024through}. This reductive approach overlooks the nuanced legal argumentation underpinning each decision, focusing on a single outcome instead of capturing the depth of legal reasoning and debate.

\noindent \textbf{Limits of Classification Benchmarks:}
The working assumption of these approaches is that by getting better at the benchmark, models encode more legal knowledge which can be extracted as explanations for predictions. To the best of our understanding, however, this promise has not been fulfilled. Initial works on data from the EtCHR, \citealt{aletras2016predicting,chalkidis2019neural}  listed words based on feature importance or highlighted  text based on  attention scores. In later works, \citet{chalkidis2021paragraph} used regularization techniques to identify paragraphs that support a finding of a violation of ECtHR. The extracted rationales did not correspond well to the annotation by a single legal expert. \citet{santosh2022deconfounding, malik2021ildc} continued the trend of computing paragraph level importance using interpretability techniques such as Integrated Gradient and tried to assess them against  expert-annotated important paragraphs, also with only moderate success. In the ECtHR context, \citet{santosh2022deconfounding} discovered evidence that BERT-based classifiers rely on shallow predictors. This can be mitigated using adversarial training, but alignment still remains low. Recently, \citet{xu2023dissonance} assessed rationale alignment at the more difficult, fine-grained word level. The experiment uncovered inconsistencies in the court metadata and illustrated how even annotations by two legal experts may not align well. To add to the challenge, a pilot study by \citet{branting2021scalable} discovered that human performance in a prediction task does not improve if users are given access to a saliency map derived from a prediction model. Recent work by \citet{mumford2023human} reported that human performance on the judgment prediction task closely resembled randomness and was unaffected by domain knowledge. These results all cast doubt on the assumption that, at least for classifiers models, benchmark performance correlates with better explanations. The data may be noisy, the labeling too simplified, the predictors too shallow, the expert disagreement low, and the utility of a salience map limited. It should also be noted that the potential leakage of benchmark test data into training corpora remains under-discussed and unmeasured.

Other body of works on outcome classification of ECtHR cases predict the decision from a textual description of the case facts alone. By contrast, what lawyers actually need is the explanation why the resolution of a case is the proper application of the law and in line with what traditional AI \& Law work would call a `theory' of ECtHR jurisprudence. The outcome must be based on a justification which presents equitable arguments, can be reviewed on appeal, and hold up under public scrutiny. 


\noindent \textbf{Shift to Generative Models:}
LLMs have also been evaluated against case outcome classification as a benchmark.  \citet{chalkidis2023chatgpt,vats2023llms,trautmann2022legal,shui2023comprehensive} tested various early models and found them to score relatively low in quantitative metrics, which stands in contrast to their scores on some bar exams \cite{katz2023gpt,freitas2023does}.
They report on experiments with several models and prompting techniques, including zero/few-shot prompting, prompt ensembling, chain-of-thought, and activation fine-tuning. \citet{yu2022legal,yu2023exploring} employ prompts that are derived from legal reasoning methods (such as the common law IRAC (Issue, Rule, Application, Conclusion). \citet{trautmann2023large} uses prompt chaining with an initial summarization step to deal with lengthy legal documents. \citet{jiang2023legal,deng2023syllogistic} develop syllogism prompting providing the three deductive reasoning steps for major premise (article/law retrieval), minor premise (element extraction from facts) and conclusion (judgement). 

LegalBench \cite{guha2023legalbench} recently presented the first aggregated benchmark beyond classification-like evaluation to test the reasoning abilities of generative models. \citet{kang2023can} applies the IRAC methodology comprehensively to LegalBench subtasks. While ancillary challenges remain (e.g., the need to manually assess model performance non certain tasks), this development is in line with our arguments in this paper.



\section{Challenges \& Future Directions}
\noindent \textbf{Combining Knowledge and Data:} The pressing question is how best to integrate legal knowledge and ML so that a system can learn from data and still seamlessly interface to a lawyer's understanding of the domain by means of a conceptual representation. A number of such hybrid systems can be found outside of NLP: \textit{Split Up} \cite{stranieri1999hybrid}  combined expert-crafted rules and neural networks trained from data in a factor-based model of Australian family law to predict divorce asset division. In the CATO line of work, both AGATHA \cite{chorley2005agatha} and VJAP \cite{grabmair2017predicting} leveraged structured legal argumentation for prediction with signals derived from a case base. Moving to NLP, one intuitive combination is to ascribe factors from cases using text processing and proceed with formalized legal inference. This was employed in SCALE \cite{branting2021scalable} to enable a logic model to predict WIPO domain name disputes, and in the ECtHR domain by inference using an ADF representation   \citet{mumford2022reasoning,mumford2023combining}. 
\citet{gray2023automatic} automatically identified factors in Fourth Amendment auto stop cases, demonstrated their predictive value, and used ML techniques 
to explain case outcomes in terms legal professionals can understand.
\citet{holzenberger2021factoring} apply neural models to identify argument slots in legal provisions and find suitable filling elements from fact descriptions, thereby enabling rule-based inference. Similarly, \citet{holzenberger2023connecting} automates the translation of cases into a knowledge base by posing it as an information extraction task. 

\noindent \textbf{Data Utilized:} Ideally, outcome prediction systems in the legal domain should rely on the information available before proceedings start and legal conclusion are determined (e.g., argumentative memoranda from the parties). Most case outcome classification research is conducted based on fact descriptions that are taken from judgments. These are often highly selective summaries tailored to align with the decision \cite{tippett2021does}. Although they may not explicitly contain outcomes, this can introduce confounding effects as demonstrated in \citet{santosh2022deconfounding}. To illustrate the effect of proxy data on performance, \citet{medvedeva2021automatic} utilized data from ECtHR `communicated cases', court-prepared summary data derived from applicant submissions, published before trial and observed a decline compared to facts statements from judgments, highlighting the need to select appropriate data for this task to draw reliable conclusions
\cite{medvedeva2023rethinking,medvedeva2023legal}. Such work may also be subject to data selection bias related to which cases reach which court, and with regard to how they are published. For example, a higher court will receive a different distribution of cases (i.e., such with grounds for appeal) than a district court, and only a subset of them may be published.
Finally, many cases are settled before or during trial, further skewing the dataset \cite{ osbeck2018outcome}. 

\noindent \textbf{Temporal Dynamics} 
Current legal NLP methods often operate under the implicit assumption that past training data is homogeneous and neglect its sequential nature. In reality, attitudes and case law change over time, with later cases altering and superseding the roles of older ones. 
All shifts in jurisprudence confront the model with a cold start problem of little training data for a new legal rule and copious training data for outdated ones. 
These dynamics can in principle be modeled. For instance, overruling detection can identify where previous legal precedents have been overturned, 
and trigger techniques such as model unlearning (see \citealt{nguyen2022survey}) or selective forgetting. 
One can also strive to detect updates in beliefs/knowledge expressed in decisions over time, and modify such beliefs within the model \cite{hase2021language}. \citet{santosh2024chronoslex} accounts for the temporally evolving nature of classification tasks on legal data using continual learning approaches.
Overall, however, the temporal dynamics of legal corpora remain largely unaddressed in recent works.

\noindent \textbf{Domain Model Construction}
Rule-based models of the law are powerful tools to develop software that supports legal practice, but constructing them demands considerable legal expertise.
Modern LLMs put us into a position to create these structures in a (semi-) automated fashion. \citet{savelka2023can} shows constructive evidence of this, but it remains an open questions whether LLMs can systematize large complexes of legal source material into well-formed, legally correct representations. 
Ascribing factors from facts text in unseen cases by means of developing classifiers requires training data relative to an exhaustively defined list of factors. The more likely scenario is that generative models can be prompted with specific facts to subsume them under a factor pattern description. For example, \citet{gray2024empirical} applied generative AI automatically to identify factors in Fourth Amendment auto stop cases.

\noindent \textbf{From Argument Mining to Generation}:
The task of constructing abstract argumentation models closely dovetails with the field of argument mining (i.e., the detection of argumentative text segments and their interlinking). Traditionally, argument mining mainly encompasses four sub-tasks as formalized by seminal work in \citealt{palau2009argumentation}: text segmentation, argument span detection, classification (e.g., conclusion, premise), and prediction of graph relations between spans. Follow up work by \citealt{wyner2010approaches,grabmair2015introducing,poudyal2020echr,habernal2023mining,ali2022constructing,ali2023legal,grundler2022detecting} focused on the first three subtasks, with fewer models engaging in graph construction. Modeling the relationships and comparative strength between conflicting arguments is a crucial piece to connect these extractive argumentative mining efforts to structured argumentation, largely unaddressed by existing works.

Even with powerful LLMs available, optimal argumentation support systems for legal practitioners benefit from structured representations of legal information and argumentation. While argumentative text can now be generated by current models, it remains a challenging cognitive task to systematize and assess arguments strategically. A productive support system should produce arguments in a transparent manner, and offer the user an intuitive way of resolving multiple complex arguments towards a justification of a decision. Naturally, this also entails questions around mindful interface design and organizational processes to facilitate accountable human decision making where capable text generation systems are accessible. 

\noindent \textbf{Role of Evaluation:} The true value in NLP for legal applications lies in producing, structuring, and assessing arguments about legal conclusions in an explainable way so that they may maximally support human experts. This human-centric nature of legal systems introduces a level of complexity that purely data-driven systems often struggle to capture when classifying variables from close-to-raw data. By the same token, LLMs may generate text that reads lawyer-like, but integrating them in processes of legal practice regularly involves interfacing them with symbolic data structures on both input and output ends, as well as maximizing consistency and correctness of generated text in ways that is defined by the legal concepts of the application context. This may include obfuscating cumbersome and error-prone model prompting behind traditional user interfaces composed of elements that map to symbols in the domain (e.g., types of contract clauses, factor-like aspects of cases, information elements of interest to draft process memoranda, etc.). The complexity of human legal decision-making highlights the inadequacy of current evaluation metrics. Legal NLP works should, ideally, tangibly indicate progress towards optimal argumentation support systems for legal practitioners, yet frequently convenient evaluations are prioritized over informative ones. This is, of course, due to the nuanced and often ill-defined characteristic of legal practice tasks. Still, legal databases are more than large repositories of text for autoregressive pre-training, but resources tend towards tackling these use cases, including, for example, using prior decisions in constructing and responding to arguments. Legal NLP’s efforts should be evaluated - and reviewed - in terms of how well models provide such functionality \cite{ashley2022prospects}. Many legal NLP works specify use cases, yet few account for them in their evaluative framework by conducting studies with legal experts, or benchmark their automatic metrics against human evaluations. Research on evaluation criteria that better capture the practical utility of legal NLP systems in real-world settings should be among our top priorities.

Examples of human evaluations in specified use cases include the following: In \citet{elaraby2024adding} human experts  evaluated the legal argument coverage in generated summaries. In \citet{mullick2022evaluation} and \citet{salaun2022conditional}, humans assessed legal summaries’ relevance, readability, fluency, or adequacy. In \citet{xu2022multi} expert evaluators  assessed the information quality of legal summaries in terms of generated question-answer pairs. Experts evaluated the legal importance of automatically identified paragraphs in 
\citet{santosh2022deconfounding} but achieving expert annotation agreements is challenging, especially given noisy metadata \cite{xu2023dissonance}. 
Evaluations benchmarked human classification of case verdicts under ECHR Article 6 in \citet{mumford2023human} and compared
expert annotations to automatically generated explanations in \citet{malik2021ildc} and to automatically identified factor sentences in \citet{gray2023automatic}.

\section{Conclusions}
We believe that knowledge-based approaches to building legal argument support systems deserve the attention of the modern NLP community, as they embody a culture and method of capturing intricacies of legal systems and argumentation that are often simplified away in the increasingly easier application of large mainstream models to legal data. The prominent role of benchmarks compounds this by drawing attention towards quantitative progress instead of real, empirical investigations of downstream benefit to practitioners. At the same time, LLMs widen the knowledge acquisition bottleneck for structured models considerably, opening up new opportunities. We believe there is great value in combining knowledge- and data-driven systems rather than continuing the assumption that deep expertise will reliably emerge given large enough amounts of data and computation.

\section*{Limitations}
This paper focuses on legal NLP as applied to tasks that involve the application of legal source material to case facts, analysis of case texts, and legal argumentation in general. Other subfields of NLP in the legal domain do not focus on argumentation about the legal significance of case facts, such as technology-assisted review in e-Discovery, contract analysis, and patent search. Similarly, legal question answering, automatic summarization of judgments, legal information retrieval, and models supporting regulatory compliance, although important, are in focus for our argumentation-related narrative. We strive to synthesize a very broad notion of the important role of expert legal knowledge to facilitate better NLP systems that will be of high utility to the stakeholders involved in the ecosystem. In our way, the way forward requires input from diverse perspectives and collaboration across multiple disciplines, including law, computer science, linguistics, and ethics to achieve a comprehensive understanding of the challenges and opportunities. We hope that the insights provided in this paper will stimulate an open discussion within the legal NLP community and beyond. 

\section*{Ethics Statement}
It is important to acknowledge that utilizing historical data to train data-driven models may inadvertently introduce biases into the system. For example, \citet{chalkidis2022fairlex} investigated disparities in classification performance based on factors such as gender, age, and respondent state in human rights litigation. Similar efforts to scrutinize for fairness and bias have been undertaken by \citet{wang2021equality,santosh2024towards,li2022fairness}. Moreover, recent pre-trained models can inherit biases encoded within their pre-training data. Therefore, any data-driven legal NLP system intended for practical deployment must undergo rigorous scrutiny to ensure compliance with applicable equal treatment and transparency imperatives. This should encompass their performance, behavior, and intended application.

We reiterate the pioneering work in AI \& Law by \citealt{buchanan1970some}, which suggested that the computer modeling of legal reasoning would be a fruitful area for research, so as to foster better understanding of legal reasoning and legal argument formation. While we do not advocate for the direct application of predictive systems within courts, the contributions of this paper are intended to facilitate research in this area to enhance transparency, accountability, and explainability. Our goal is to align NLP systems supporting legal practitioners as closely as possible with legal expertise, and to contribute to the discussion around their ethical use.

\bibliography{custom}
\bibliographystyle{acl_natbib}

\appendix


\end{document}